\pdfoutput=1

\documentclass[11pt]{article}

\usepackage[]{acl}

\usepackage{times}
\usepackage{latexsym}
\usepackage{times}
\usepackage{multirow}
\usepackage{multicol}
\usepackage{graphicx}
\usepackage{amsmath}
\usepackage{amssymb}
\usepackage{bbm}
\usepackage{subfigure}
\usepackage{amsfonts}
\usepackage{bm}
\usepackage{bbding}
\usepackage{booktabs}
\usepackage{color}
\usepackage[T1]{fontenc}

\usepackage[utf8]{inputenc}
\usepackage{microtype}
\usepackage{CJKutf8}
\usepackage{makecell}

\title{SEE-Few: Seed, Expand and Entail for Few-shot Named Entity Recognition}

\author{Zeng Yang \and Linhai Zhang \and Deyu Zhou\thanks{\ \  Corresponding author.} \\
School of Computer Science and Engineering, Key Laboratory of Computer Network \\ and Information Integration, Ministry of Education, Southeast University, China \\
\texttt{\{yangzeng, lzhang472, d.zhou\}@seu.edu.cn}}

\begin{document}
\maketitle

\begin{abstract}
Few-shot named entity recognition (NER) aims at identifying named entities based on only few labeled instances. Current few-shot NER methods focus on leveraging existing datasets in the rich-resource domains which might fail in a training-from-scratch setting where no source-domain data is used. To tackle training-from-scratch setting, it is crucial to make full use of the annotation information (the boundaries and entity types). Therefore, in this paper, we propose a novel multi-task (Seed, Expand and Entail) learning framework, SEE-Few, for Few-shot NER without using source domain data. The seeding and expanding modules are responsible for providing as accurate candidate spans as possible for the entailing module. The entailing module reformulates span classification as a textual entailment task, leveraging both the contextual clues and entity type information. All the three modules share the same text encoder and are jointly learned. Experimental results on four benchmark datasets under the training-from-scratch setting show that the proposed method outperformed state-of-the-art few-shot NER methods with a large margin. Our code is available at \url{https://github.com/unveiled-the-red-hat/SEE-Few}. 

\end{abstract}

\section{Introduction}

Named entity recognition (NER), focusing on identifying mention spans in text inputs and classifying them into the pre-defined entity categories, is a fundamental task in natural language processing and widely used in downstream tasks~\cite{wang-etal-2019-open,ZHOU2021107200,9428547}. Supervised NER has been intensively studied and yielded significant progress, especially with the aid of pre-trained language models~\cite{devlin2019bert,li_unified_2020,mengge_coarse--fine_2020,yu_named_2020,shen2021locateandlabel,li_modularized_2021,chen_enhancing_2021}. However, supervised NER relies on plenty of training data, which is not suitable for some specific situations with few training data. 

Few-shot NER, aiming at recognizing entities based on few labeled instances, has attracted much attention in the research filed. Approaches for few-shot NER can be roughly divided into two categories, span-based and sequence-labeling-based methods. Span-based approaches enumerate text spans in input texts and classify each span based on its corresponding template score~\cite{cui2021templatebased}. 
Sequence-labeling-based approaches treat NER as a sequence labeling problem which assigns a tag for each token using the BIO or IO tagging scheme~\cite{yang2020simple,hou2020few,huang2020few}. Most of these span-based and sequence-labeling-based methods focus on leveraging existing datasets in the rich-resource domains to improve their performance in the low-resource domains. Unfortunately, the gap between the source domains and the target domains may hinder the performance of these methods~\cite{pan2009transfer_survey,cui2021templatebased}. Moreover, these approaches might fail under the training-from-scratch setting where no source domain data is available.

Therefore, it is crucial to make full use of the in-domain annotations, which consist of two types of information: boundary information and entity type information. However, most the approaches mentioned above fail to fully utilize these information. (1) Most span-based methods simply enumerate all possible spans, ignoring the boundary information of named entities. As a large number of negative spans are generated, these approaches suffer from the bias, the tendency to classify named entities as non-entities. (2) Most sequence-labeling-based methods simply employ one-hot vectors to represent entity types while ignoring the prior knowledge of entity types. 

To overcome the disadvantages mentioned above, firstly, inspired by three principles for weakly-supervised image segmentation, i.e. seed, expand and constrain~\cite{kolesnikov2016seed}, we seed with relatively high-quality unigrams and bigrams in the texts, then expand them to extract the candidate spans as accurately as possible. Secondly, we cast span classification as textual entailment to naturally incorporate the entity type information. For example, to determine whether ``J. K. Rowling'' in ``J. K. Rowling is a British author.'' is a {\tt PERSON} entity or a non-entity, we treat ``J. K. Rowling is a British author.'' as a premise, then construct ``J. K. Rowling is a person.'' and ``J. K. Rowling is not an entity.'' as hypotheses. In such way, span classification is converted into determining which hypothesis is true. Moreover, the size of training data is increased by such converting which is beneficial for few-shot settings. 

In this paper, we propose SEE-Few, a novel multi-task learning framework (Seed, Expand and Entail) for Few-shot NER. The seeding and expanding modules are responsible for providing as accurate candidate spans as possible for the entailing module. Specifically, the seed selector chooses some unigrams and bigrams as seeds based on some metrics, e.g., the Intersection over Foreground. The expanding module takes a seed and the window around it into account and expands it to a candidate span. Compared with enumerating all possible {\^n}-gram spans, seeding and expanding can significantly reduce the number of candidate spans and alleviate the impact of negative spans in the subsequent span classification stage. The entailing module reformulates a span classification task as a textual entailment task, leveraging contextual clues and entity type information to determine whether a candidate span is an entity and what type of entity it is. All the three modules share the same text encoder and are jointly learned. Experiments were conducted on four NER datasets under training-from-scratch few-shot setting. Experimental results show that the proposed approach outperforms several state-of-the-art baselines.

The main contributions can be summarized as follows:
\begin{itemize}
\item 
A novel multi-task learning framework (Seed, Expand and Entail), SEE-Few, is proposed for few-shot NER without using source domain data. In specific, the seeding and expanding modules provide as accurate candidate spans as possible for the entailing module. The entailing module reformulates span classification as a textual entailment task, leveraging contextual clues and entity type information.
\item Experiments were conducted on four NER datasets in training-from-scratch few-shot setting. Experimental results show that the proposed approach outperforms the state-of-the-art baselines by significant margins. 
\end{itemize}

\begin{figure*}[htp]
    \centering{\includegraphics[width=\linewidth]{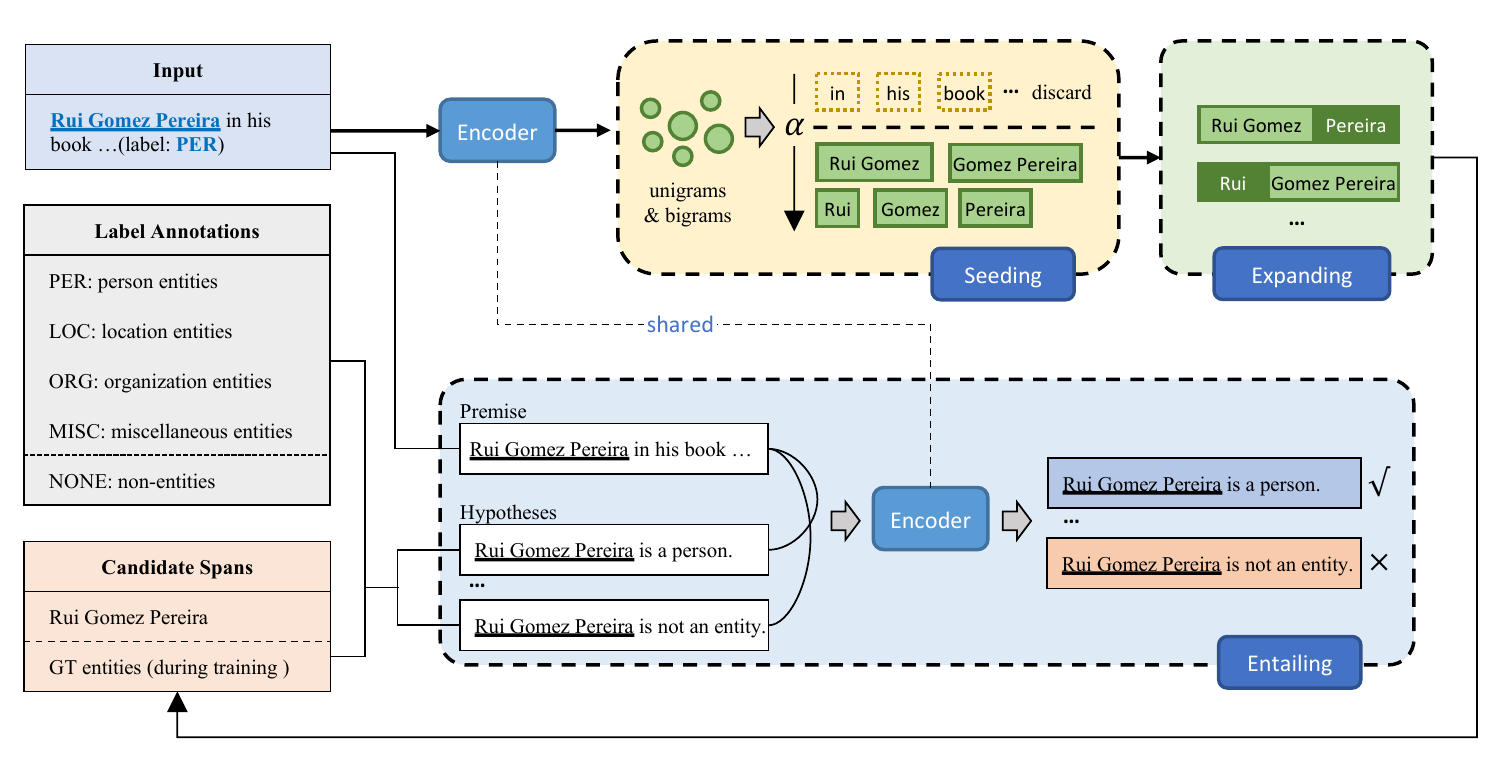}}
    \caption{The architecture of the proposed approach, SEE-Few, which consists of three main modules: seeding, expanding, and entailing.}
    \label{fig:model}
\end{figure*}

\section{Related Work}

\subsection{Few-shot NER}

Few-shot NER aims at recognizing entities based on only few labeled instances from each category. A few approaches have been proposed for few-shot NER. Methods based on prototypical network \cite{snell2017prototypical} require complex episode training~\cite{fritzler2019few,hou2020few}. \citet{yang2020simple} abandon the complex meta-training and propose NNShot, a distance-based method with a simple nearest neighbor classifier. 
\citet{huang2020few} investigate three orthogonal schemes to improve the model generalization ability for few-shot NER. 
TemplateNER~\cite{cui2021templatebased} enumerates all possible text spans in input text as candidate spans and classifies each span based on its corresponding template score. 
\citet{ma2021templatefree} propose a template-free method to reformulate NER tasks as language modeling (LM) problems without any templates. \citet{tong2021learning} propose to mine the undefined classes from miscellaneous other-class words, which also benefits few-shot NER. \citet{ding-etal-2021-nerd} present Few-NERD, a large-scale human-annotated few-shot NER dataset to facilitate the research. 

However, most of these studies follow the manner of episode training~\cite{fritzler2019few,hou2020few,tong2021learning,ding-etal-2021-nerd} or assume an available rich-resource source domain~\cite{yang2020simple,cui2021templatebased}, which is in contrast to the real word application scenarios that only very limited labeled data is available for training and validation~\cite{ma2021templatefree}. EntLM~\cite{ma2021templatefree} is implemented on training-from-scratch few-shot setting, but still needs distant supervision datasets for label word searching. The construction of distant supervision datasets requires additional expert knowledge. Some works study generating NER datasets automatically to reduce labeling costs~\cite{simple-gen-dataset,logical-rule}. In this paper, we focus on the few-shot setting without source domain data which makes minimal assumptions about available resources.

\subsection{Three Principles for Weakly-Supervised Image Segmentation}

Semantic image segmentation is a computer vision technique which aims at assigning a semantic class label to each pixel of an image. \citet{kolesnikov2016seed} introduce three guiding principles for weakly-supervised semantic image segmentation: to seed with weak localization cues, to expand objects based on the information of possible classes in the image, and to constrain the segmentation with object boundaries. 

\section{Methodologies}

\subsection{Problem Setting}

We decompose NER to two subtasks: span extraction and span classification. Given an input text $\mathbf{X}=\{x_1,\dots,x_n\}$ as a sequence of tokens, a span starting from $x_l$ and ending with $x_r$ (i.e., $\{x_l,\dots,x_r\}$) can be denote as $s=(l,r)$, where $1 \leq l \leq r \leq n$. The span extraction task is to obtain a candidate span set $\mathbf{C}=\{c_1,\dots,c_{m}\}$ from the input text. Given an entity type set $\mathbf{T}^+=\{t_1,\dots,t_{v-1}\}$ and the candidate span set $\mathbf{C}$ produced by span extraction, the target of span classification is assign an entity category $t \in \mathbf{T}^+$ or the non-entity category to each candidate span. For convenience, we denote an entity type set including the non-entity type as $\mathbf{T}=\{t_1,\dots,t_{v-1},t_{none}\}$, where $t_{none}$ represents the non-entity type and $v$ is the size of $\mathbf{T}$.

\subsection{The Architecture}
Figure~\ref{fig:model} illustrates the architecture of the proposed approach, SEE-Few, which consists of three main modules: seeding, expanding, and entailing. The input text will first be sent to the seeding module to generate informative seeds, then the seeds will be expanded to candidate spans in the expanding module, finally the candidate spans will be classified with an entailment task in the entailing module. We will discuss the details of each modules in the following sections. 

\subsubsection{Seeding}
Given an input text $\mathbf{X}=\{x_1,\dots,x_n\}$ consisting of $n$ tokens, a unigram consists of one token and a bigram consists of two consecutive tokens. We denote the set of unigrams and bigrams in the input text as $\mathbf{S}=\{s_1,\dots,s_{2n-1}\}$, where $s_i=(l_i,r_i)$ denotes $i$-th span, and $l_i$, $r_i$ denote the left and right boundaries of the span respectively. 

Seeding is to find the unigrams and bigrams that overlap with entities and have the potential to be expanded to named entities, which is important for the following seed expansion. It can be accomplished by constructing a seeding model and predicting the seed score for each candidate unigram or bigram. 

Firstly, we feed the input text into BERT to obtain the representation $h \in \mathbb{R}^{n \times d}$, where $d$ is the dimension of the BERT hidden states. For the span $s_i=(l_i,r_i)$, its representation $h_i^{seed}$ is the concatenation of the mean pooled span representation $h_i^p$ and the representation of the \texttt{[CLS]} token $h^{\tt [CLS]}$. The seed score is calculated as follows:

\begin{equation}
    h^p_{i} = \operatorname{MeanPooling}(h_{l_{i}},\dots,h_{r_{i}})
\end{equation}
\begin{equation}
    h_i^{seed} = \operatorname{Concat}(h^p_{i}, h^{\tt [CLS]})
\end{equation}
\begin{equation}
    p_i^{seed} = \operatorname{Sigmoid}(\operatorname{MLP_s}(h_i^{seed}))
\end{equation}

\noindent where MLP denotes the multilayer perceptron with a GELU function in the last layer. We set the threshold $\alpha$ and select the span whose seed score is above $\alpha$ as a seed to expand. 

To train the seeding model, we need to construct a dataset consisting of unigrams (bigrams) and their seeding scores. We construct the seed score based on Intersection over Foreground (IoF). Intersection over Union (IoU) is used to measure the overlap between objects in object detection which is defined as $\operatorname{IoU}(A, B) =\frac{A\cap B}{A\cup B}$ in NER, where $A$ and $B$ are two spans~\cite{chen2020boundary,shen2021locateandlabel}. However, IoU is not suitable for the seeding stage. Considering an entity consisting of five words, e.g., ``International Conference on Computational Linguistics'', IoU between the bigram ``International Conference'' and the entity is 0.4, not significant. Intersection over Foreground (IoF) can be a better choice which is defined as $\operatorname{IoF}(A, B) =\frac{A\cap B}{A}$, where $A$ is the foreground (i.e., a unigram or bigram) and $B$ is the background (i.e., a named entity). In the above example, IoF between ``International Conference'' and the entity is 1.0, indicating that it is part of the entity and has the potential to be expanded to the whole entity. We assign each $s$ the IoF between it and its closed named entity as the ground-truth seed score $\hat{y}_i^{seed}$ which indicates the potential to be expanded to the whole entity.

\subsubsection{Expanding}

For named entities consisting of more than two words, the seeds generated in seeding stage are only part of them and need to be expanded to the whole entities. Expanding is a regression task to learn the boundary offsets $\hat{o}$ between a seed and the named entity closed to the seed. 

Expanding is allowed to offset the left and right boundaries of the seed by up to $\lambda$, respectively, which means that the longest entity we can get is an entity of length $2+2\lambda$. Besides, expanding needs to consider a window around the seed in addition to the seed itself. For the seed $s_i = (l_i,r_i)$, the maximum expansion is denoted as:

\begin{equation}
    s_i^{exp\_max} = (\max(1, l_i-\lambda),\min(n, r_i+\lambda))
\end{equation}

If we use $s_i^{exp\_max}$ as the window around $s_i$, it may not provide enough information to distinguish the boundaries for the maximum expansion. Thus, the window for $s_i$ should be larger than $s_i^{exp\_max}$, defined as $w_i$:

\begin{equation}
    w^l_i = \max(1, l_i-2\lambda)
\end{equation}

\begin{equation}
    w^r_i = \min(n, r_i+2\lambda)
\end{equation}

\begin{equation}
    w_i = (w^l_i,w^r_i)
\end{equation}

We concatenate the mean pooled span representation $h_i^p$ of seed $s_i$ and the mean pooled span representation $h_i^w$ of window $w_i$. Then the offsets $o_i$ of left and right boundaries are calculated as follows: 

\begin{equation}
    h^w_{i} = \operatorname{MeanPooling}(h_{w^l_i},\dots,h_{w^r_i})
\end{equation}
\begin{equation}
    h_i^{exp} = \operatorname{Concat}(h^p_{i}, h^w_{i})
\end{equation}
\begin{equation}
    o_i = \lambda\cdot(2\cdot\operatorname{Sigmoid}(\operatorname{MLP_e}(h_i^{exp})) - 1)
\end{equation}

\noindent where $o_i \in \mathbb{R}^2$. The first element of $o_i$ can be denoted as $o_i^l$, indicating the offset of the seed's left boundary. Likewise, the second element  $o_i^r$ indicates the offset of the seed's right boundary, and $o_i^l, o_i^r \in [-\lambda, \lambda]$. We can obtain the result of expanding, i.e., a candidate span with the new boundaries $l'_i$ and $r'_i$:

\begin{equation}
    l'_i = \max(1, l_i + \left\lfloor o_i^{l}+\frac{1}{2}\right\rfloor)
\end{equation}
\begin{equation}
    r'_i = \min(n, r_i + \left\lfloor o_i^{r}+\frac{1}{2}\right\rfloor )
\end{equation}

The duplicate results and invalid results that $l'_i > r'_i$ are discarded. At this point, a set of candidate spans are produced for span classification. 

\subsubsection{Entailing}

The entailing module reformulates span classification as a textual entailment task, leveraging contextual clues and entity type information. To cast span classification as textual entailment, we need to construct textual entailment pairs. For the $i$-th candidate span $c_i$, the entailment pair is constructed as $(\mathbf{X}, \mathbf{E}_{i}^{j})$, where $\mathbf{E}_{i}^{j}=\{c_i,\text{is},\text{a},t_j\}$ and $t_j \in \mathbf{T}$. Please refer to Appendix~\ref{app:template} for detailed templates used to construct entailment pairs. The entailment label $\hat{y}^{entail}_{i,j}$ for $(\mathbf{X}, \mathbf{E}_{i}^{j})$ can be obtained by:

\begin{equation}
    \hat{y}^{entail}_{i,j} = 
    \left\{
        \begin{array}{lr}
            \tt{entail}, & \text{if\ } c_i \text{ belongs to } t_j \\
            
            \tt{not \ entail}, & \text{otherwise} \\
            \end{array}
    \right.
\end{equation}

The entailment pair $(\mathbf{X}, \mathbf{E}_{i}^{j})$ is fed into the shared text encoder to obtain the representation of the \texttt{[CLS]} token $h^{\tt [CLS]}_{\mathbf{E}_{i,j}} \in \mathbb{R}^{d}$ and the binary textual entailment classification can be performed:

\begin{equation}
    p^{entail}_{i,j} = \operatorname{Softmax}(\operatorname{MLP_{entail}}(h^{\tt [CLS]}_{\mathbf{E}_{i,j}}))
\end{equation}

To ensure that all ground-truth entities are learned, we add ground-truth entities to the candidate span set $\mathbf{C}$ during the training phase.

\subsection{Training Objective}

Both seeding and expanding are regression tasks, 
the seeding loss $\mathcal{L}_{seed}$ and expansion loss $\mathcal{L}_{exp}$ are defined as follows:

\begin{equation}
\mathcal{L}_{seed}=\sum_i \operatorname{SmoothL1}\left(\hat{y}^{seed}_{i}, p^{seed}_{i}\right)
\end{equation}

\begin{equation}
\mathcal{L}_{exp}=\sum_i \sum_{j\in \{l,r\}} \operatorname{SmoothL1}\left(\hat{o}^{j}_{i}, o^{j}_{i}\right)
\end{equation}

For the entailing module, since the number of instances with the {\tt not entail} label is bigger than the number of instances with the {\tt entail} label, we use focal loss \cite{8237586} to solve the label imbalance problem: 

\begin{equation}
    FL(p,y) = 
    \left\{
        \begin{array}{lr}
            -(1-p_{i,j})^\gamma\log(p_{i,j}), & \text{if\ } y=1 \\
            
            -(p_{i,j})^\gamma  \log(1-p_{i,j}), & \text{otherwise} \\
            \end{array}
    \right.
\end{equation}

\begin{equation}
    \begin{aligned}
    \mathcal{L}_{entail} &= \sum_i \sum_j FL(p_{i,j}^{entail},\hat{y}_{i,j}^{entail})
    \end{aligned}
\end{equation}

\noindent where $\gamma$ denotes the focusing parameter of focal loss. 

The multi-task framework is trained by minimizing the combined loss defined as follows:

\begin{equation}
\mathcal{L} = \beta_{1}\mathcal{L}_{seed} + \beta_{2}\mathcal{L}_{exp} + \beta_{3}\mathcal{L}_{entail} 
\end{equation}

\noindent where $\beta_{1}$, $\beta_{2}$ and $\beta_{3}$ are hyperparameters controlling the relative contribution of the respective loss term.

\subsection{Entity Decoding}

The entailing module will output an entailment score $p^{entail}_{i,j}$ for the entailment pair $(\mathbf{X},\mathbf{E}_{i}^{j})$, where $\mathbf{E}_{i}^{j}=\{c_i,\text{is},\text{a},t_j\}$, $c_i \in \mathbf{C}$ and $t_j \in \mathbf{T}$. We collect all entailment pairs associated with the candidate span $c_i$, then assign $c_i$ the entity type with the highest entailment score. If two candidate spans have overlap, the span with a higher score will be selected as the final result.

\section{Experiment Settings}

\subsection{Training-from-scratch Few-shot Settings}

Different from most previous few-shot NER studies that assume source-domain data is available, we consider a training-from-scratch setting, which is more practical and challenging. Specifically, we assume only $\textit{K}$ examples for each entity class in the training set and validation set respectively, where $K \in \{5,10,20\}$.

\subsection{Datasets Construction}

For fair comparison, we manually construct the few-shot datasets. 
With $K \in \{5,10,20\}$, we follow the greedy sampling strategy in \cite{yang2020simple} to ensure the sample number $K$ of each category. To make the experimental results more convincing and credible, we randomly sample 5 different groups of training sets and validation sets for each $K$. We employ these strategies on four NER datasets from different domains: CoNLL2003 dataset \cite{sang2003introduction} in news domain, MIT-Restaurant dataset \cite{liu2013asgard} in review domain, WikiGold dataset \cite{balasuriya2009named} in general domain and Weibo dataset \cite{HeS16} in social media domain. Table~\ref{tab:dataset} shows the statistics on these original datasets. The self-constructed datasets are public available with the code for reproducibility. 

\begin{table*}[ht]
    \centering
    \resizebox{\linewidth}{!}{
    \begin{tabular}{c|c|rrl|rrl|rrl}
    \hline
    \hline
    \multicolumn{1}{c|}{\multirow{2}{*}{\textbf{Datasets}}} &
    \multicolumn{1}{c|}{\multirow{2}{*}{\textbf{Methods}}} &
    \multicolumn{3}{c|}{$K=5$} & \multicolumn{3}{c|}{$K=10$} & \multicolumn{3}{c}{$K=20$} 
    \\ \cline{3-11}
    \multicolumn{1}{c|}{} & \multicolumn{1}{c|}{} & \multicolumn{1}{c}{$P$}& 
    \multicolumn{1}{c}{$R$} &
    \multicolumn{1}{c|}{$F_{1}$} & 
    \multicolumn{1}{c}{$P$}& 
    \multicolumn{1}{c}{$R$} &
    \multicolumn{1}{c|}{$F_{1}$} & 
    \multicolumn{1}{c}{$P$}& 
    \multicolumn{1}{c}{$R$} &
    \multicolumn{1}{c}{$F_{1}$} 
    \\
    \hline
    \multirow{6}{*}{\textbf{CoNLL03}}&
    
    LC-BERT & 42.83 & 30.72 & $35.06_{(6.09)}$ & 50.36 & 52.13 & $51.20_{(6.39)}$ & 56.33 & 63.85 & $59.84_{(1.43)}$ \\
    & Prototype & 38.26 & 43.14 & $40.37_{(8.06)}$ & 45.08 & 64.02 & $52.82_{(3.22)}$ & 43.94 & 69.72 & $53.89_{(1.95)}$ \\
    & NNShot & 32.11 & 38.42 & $34.92_{(3.30)}$ & 34.10 & 40.98 & $37.18_{(5.82)}$ & 38.43 & 47.85 & $42.61_{(2.23)}$ \\
    & StructShot & 30.04 & 21.33 & $23.43_{(4.52)}$ & 38.62 & 19.72 & $26.09_{(7.23)}$ & 44.96 & 28.59 & $34.87_{(1.30)}$ \\
    & TemplateNER & 26.90 & 23.46 & $23.13_{(8.40)}$ & 44.51 & 43.99 & $44.01_{(4.82)}$ & 52.16 & 56.46 & $54.01_{(5.09)}$ \\
    & Ours & 60.45 & 51.27 & $\textbf{55.21}_{(3.93)}$ & 66.19 & 58.68 & $\textbf{61.99}_{(1.73)}$ & 69.49 & 67.07 & $\textbf{68.21}_{(2.60)}$ \\
    \hline
    \hline
    
    \multirow{6}{*}{\textbf{MIT-Restaurant}}&
    
    LC-BERT & 41.21 & 38.65 & $39.88_{(3.79)}$ & 43.60 & 48.93 & $46.08_{(3.75)}$ & 56.24 & 60.04 & $58.07_{(1.50)}$ \\
    & Prototype & 27.77 & 46.79 & $34.84_{(1.63)}$ & 30.37 & 50.64 & $37.97_{(2.29)}$ & 37.91 & 59.31 & $46.25_{(1.62)}$ \\
    & NNShot & 28.15 & 34.81 & $31.11_{(2.30)}$ & 30.28 & 37.65 & $33.56_{(1.48)}$ & 36.72 & 45.55 & $40.66_{(1.26)}$ \\
    & StructShot & 45.13 & 25.00 & $31.93_{(4.32)}$ & 43.94 & 28.19 & $34.30_{(2.56)}$ & 52.08 & 36.18 & $42.69_{(1.12)}$ \\
    & TemplateNER & 23.11 & 20.78 & $21.53_{(4.66)}$ & 39.45 & 28.77 & $32.71_{(8.14)}$ & 46.93 & 37.00 & $41.26_{(6.80)}$ \\
    & Ours & 53.08 & 39.47 & $\textbf{45.25}_{(3.18)}$ & 57.19 & 46.41 & $\textbf{51.20}_{(1.48)}$ & 64.79 & 57.22 & $\textbf{60.75}_{(2.07)}$ \\
    
    \hline
    \hline
    
    \multirow{6}{*}{\textbf{WikiGold}}&
    
    LC-BERT & 36.02 & 8.02 & $12.57_{(7.81)}$ & 43.13 & 8.95 & $37.72_{(7.20)}$ & 50.68 & 50.73 & $50.68_{(5.94)}$ \\
    & Prototype & 20.55 & 21.46 & $19.28_{(8.12)}$ & 23.31 & 45.21 & $30.59_{(3.95)}$ & 27.31 & 56.22 & $36.56_{(8.65)}$ \\
    & NNShot & 27.81 & 34.16 & $30.63_{(1.91)}$ & 26.36 & 37.92 & $30.93_{(4.89)}$ & 28.33 & 39.07 & $32.81_{(5.41)}$ \\
    & StructShot & 49.00 & 13.37 & $20.88_{(4.61)}$ & 43.21 & 14.19 & $21.28_{(2.96)}$ & 43.51 & 15.94 & $23.16_{(2.18)}$ \\
    & TemplateNER & 18.45 & 19.45 & $17.26_{(12.73)}$ & 38.33 & 45.37 & $41.04_{(13.19)}$ & 57.39 & 56.00 & $56.60_{(3.22)}$ \\
    & Ours & 61.23 & 41.01 & $\textbf{48.87}_{(8.01)}$ & 63.36 & 48.74 & $\textbf{54.98}_{(3.24)}$ & 69.06 & 58.25 & $\textbf{63.19}_{(1.28)}$ \\
    
    \hline
    \hline
    
    \multirow{6}{*}{\textbf{Weibo}}&
    LC-BERT & 36.93 & 26.32 & $29.95_{(13.93)}$ & 46.49 & 53.19 & $49.54_{(3.96)}$ & 54.27 & 58.53 & $56.23_{(1.48)}$ \\
    & Prototype & 14.32 & 37.68 & $20.64_{(7.07)}$ & 21.27 & 59.42 & $31.25_{(2.64)}$ & 21.27 & 59.42 & $37.39_{(2.58)}$ \\
    & NNShot & 4.64 & 10.57 & $06.45_{(2.65)}$ & 6.58 & 13.73 & $08.90_{(1.27)}$ & 11.77 & 26.61 & $16.32_{(0.80)}$ \\
    & StructShot & 16.77 & 1.53 & $02.80_{(1.63)}$ & 38.48 & 3.21 & $05.91_{(1.93)}$ & 52.05 & 5.93 & $10.65_{(1.73)}$ \\
    & TemplateNER & 4.12 & 16.70 & $04.41_{(4.67)}$ & 5.12 & 27.27 & $08.31_{(3.11)}$ & 10.70 & 29.57 & $15.24_{(7.09)}$ \\
    & Ours & 49.51 & 48.51 & $\textbf{48.67}_{(4.05)}$ & 55.12 & 57.65 & $\textbf{56.07}_{(1.62)}$ & 57.10 & 57.70 & $\textbf{57.21}_{(1.62)}$ \\
    
    \hline
    \hline
    
    \end{tabular}}
    \caption{Performance comparison of SEE-Few and baselines on four datasets under different $K$s. }
    \label{tab:main_result}
    \end{table*}

\begin{table}
    \centering
    
    \resizebox{\linewidth}{!}{
    \begin{tabular}{lcccrr}
    \toprule
    \textbf{Dataset} & \textbf{Domain} & \textbf{Language} & \textbf{\# Class} & \textbf{\# Train} & \textbf{\# Test} \\
    \midrule
    CoNLL03 & News & English & 4 & 14,987 & 3,684 \\
    MIT-Restaurant & Review & English & 8 & 7,660 & 1,521 \\
    WikiGold & General & English & 4 & 1,017 & 339 \\
    Weibo & Social Media & Chinese & 8 & 1,350 & 270 \\
    \bottomrule
    
    \end{tabular}}
    \caption{Statistics on the original datasets used to construct our few-shot datasets.}
    \label{tab:dataset}
    \end{table}

\subsection{Baselines}

We compare the proposed model with five competitive baselines.

\textbf{LC-BERT} \cite{devlin2019bert} BERT with a linear classifier which is applied to project the contextualized representation of each token into the label space. 

\textbf{Prototype} \cite{huang2020few} A method based on prototypical network \cite{snell2017prototypical}, represents the entity categories as vectors in the same representation space of individual tokens and utilizes the nearest neighbor criterion to assign the entity category.

\textbf{NNShot} and \textbf{StructShot} \cite{yang2020simple} NNShot is a metric-based few-shot NER method that leverages a nearest neighbor classifier for few-shot prediction. StructShot is based on NNShot and use the Viterbi algorithm for decoding predictions. These methods pre-train the model with a dataset from other rich-resource domain (source domain) which is unavailable in our training-from-scratch setting. We re-implement them and directly apply them on target domains.

\textbf{TemplateNER} \cite{cui2021templatebased} A template-based prompt learning method which fine-tunes BART \cite{lewis-etal-2020-bart} to generate pre-defined templates filled by enumerating text spans from input texts.

\begin{table*}[ht]
    \centering
    \resizebox{\linewidth}{!}{
    \begin{tabular}{l|rrr|rrr|rrr|rrr}
    \toprule
    \multirow{2}{*}{Model} & \multicolumn{3}{c|}{CoNLL03} & \multicolumn{3}{c|}{MIT-Restaurant} & \multicolumn{3}{c|}{WikiGold} & \multicolumn{3}{c}{Weibo}\\
    \cmidrule(lr){2-13}
    & \multicolumn{1}{c}{$P$} & \multicolumn{1}{c}{$R$} & \multicolumn{1}{c|}{$F_1$} & \multicolumn{1}{c}{$P$} & \multicolumn{1}{c}{$R$} & \multicolumn{1}{c|}{$F_1$}& \multicolumn{1}{c}{$P$} & \multicolumn{1}{c}{$R$} & \multicolumn{1}{c|}{$F_1$} & \multicolumn{1}{c}{$P$} & \multicolumn{1}{c}{$R$} & \multicolumn{1}{c}{$F_1$}  \\
    \cmidrule(lr){1-13}
    Full model & 60.45 & 51.27 & $\textbf{55.21}$ & 53.08 & 39.47 & $\textbf{45.25}$ & 61.23 & 41.01 & $\textbf{48.87}$ & 49.51 & 48.51 & $\textbf{48.67}$ \\
    \quad w/o seeding &  63.20 & 45.46 & 52.36 & 53.45 & 38.11 & 44.15 & 61.79 & 36.66 & 45.68 & 56.46 & 40.72 & 47.23 \\ 
    \quad w/o expanding  & 61.70 & 43.41 & 50.66 & 52.13 & 34.30 & 41.27 & 61.93 & 27.48 & 37.58 & 48.62 & 28.13 & 35.22 \\
    \quad w/o entailing  &  50.93 & 41.15 & 45.01 & 46.60 & 31.43 & 37.13 & 69.39 & 28.47 & 40.02 & 38.32 & 22.58 & 27.14 \\
    \quad w/o seed \& exp & 57.29 & 41.35 & 47.59 & 54.73 & 33.77 & 41.55 & 71.87 & 23.10 & 34.43 & 49.06 & 24.26 & 31.97 \\
    \quad repl IoF with IoU & 60.10 & 44.14 & 50.44 & 36.65 & 31.07 & 33.58 & 51.47 & 27.21 & 34.50 & 66.92 & 13.49 & 20.97 \\
    \bottomrule
    \end{tabular}}
    \caption{Ablation study on 5-shot setting with the metrics of precision, recall and F1-score. }
    \label{tab:ablation1}
    \end{table*}
    
\begin{table*}[ht]
\centering
\resizebox{\linewidth}{!}{
\begin{tabular}{l|rr|rr|rr|rr}
\toprule
\multirow{2}{*}{Model} & \multicolumn{2}{c|}{CoNLL03} & \multicolumn{2}{c|}{MIT-Restaurant} & \multicolumn{2}{c|}{WikiGold} & \multicolumn{2}{c}{Weibo}\\
\cmidrule(lr){2-9}
& \multicolumn{1}{c}{$|\mathbf{C}|$} & \multicolumn{1}{c|}{$|\mathbf{C}|\ /\ \text{\#Sen}$} & \multicolumn{1}{c}{$|\mathbf{C}|$} & \multicolumn{1}{c|}{$|\mathbf{C}|\ /\ \text{\#Sen}$} & \multicolumn{1}{c}{$|\mathbf{C}|$} & \multicolumn{1}{c|}{$|\mathbf{C}|\ /\ \text{\#Sen}$} & \multicolumn{1}{c}{$|\mathbf{C}|$} & \multicolumn{1}{c}{$|\mathbf{C}|\ /\ \text{\#Sen}$}  \\
\cmidrule(lr){1-9}
Full model & 13464 ($\times$0.15) & 3.65 & 7435 ($\times$0.28) & 4.48 & 1461 ($\times$0.11) & 4.31 & 1030 ($\times$0.04) & 3.81 \\
\quad w/o seeding & 59053 ($\times$0.66) & 16.03 & 17780 ($\times$0.66) & 15.74 & 9924 ($\times$0.72) & 29.27 & 19772 ($\times$0.67) & 73.23 \\
\quad w/o expanding  & 15055 ($\times$0.17) & 4.09 & 8281 ($\times$0.31) & 5.11 & 1662 ($\times$0.12) & 4.90 & 2093 ($\times$0.07) & 7.75 \\
\quad w/o entailing  & 19350 ($\times$0.22) & 5.25 & 7625 ($\times$0.28) & 4.68 & 1656 ($\times$0.12) & 4.88 & 1272 ($\times$0.04) & 4.71\\
\quad w/o seed \& exp & 89648 ($\times$1.00) & 24.33 & 26991 ($\times$1.00) & 17.75 & 13833 ($\times$1.00) & 40.81 & 29352 ($\times$1.00) & 108.71 \\
\quad repl IoF with IoU & 11190 ($\times$0.12) & 3.04 & 4098 ($\times$0.15) & 1.76 & 490 ($\times$0.04) & 1.45 & 276 ($\times$0.01) & 1.02\\ 

\cmidrule(lr){1-9}

\#Entity / \#Sent & \multicolumn{2}{c|}{1.53} & \multicolumn{2}{c|}{2.07} & \multicolumn{2}{c|}{2.15} & \multicolumn{2}{c}{1.55} \\

\bottomrule
\end{tabular}}
\caption{Ablation study on 5-shot setting with entity-related statistics. $|\mathbf{C}|$ denotes the number of candidate spans during the testing phase. $|\mathbf{C}|\ /\ \text{\#Sen}$ denotes the average number of candidate spans per sentence during the testing phase. \#Entity / \#Sent denotes the average number of named entities per sentence. $(\cdot)$ indicates the ratio to the number of candidate spans that produced by w/o seed \& exp (e.g., the number of all unigrams and bigrams).}
\label{tab:ablation2}
\end{table*}

\subsection{Implementation Details}

For the proposed model and all the baselines except TemplateNER, we implement them based on ``bert-base-uncased'' for English datasets and ``bert-base-chinese'' for Chinese datasets. 
TemplateNER uses BART-large~\cite{lewis-etal-2020-bart} as the backbone on English datasets and Chinese BART-large~\cite{shao2021cpt} as the backbone on Chinese datasets. 
For all the baselines, we use the recommended parameters provided by the original paper or the official implementation. 

For the proposed model, the number of epochs is 35. The batch sizes of seeding and expanding are 1, and the batch sizes of entailing are 16, 16, 16, 8 on CoNLL03, MIT-Restaurant, WikiGold and Weibo, respectively. The threshold $\alpha$s on CoNLL03, MIT-Restaurant, WikiGold and Weibo, are set to 0.5, 0.6, 0.7, 0.7, respectively. $\lambda$ is set to 5. The focusing parameter of focal loss $\gamma$ is set to 2. $\beta_1$, $\beta_2$ and $\beta_3$ are set to 1, 1 and 1, respectively. The dropouts before the seeding, expanding and entailing  are set with a rate of 0.5. The loss function is minimized using AdamW optimizer with a learning rate of 3e-05 and a linear warmup-decay learning rate schedule. 

\section{Experimental Results}

\subsection{Overall Results}

Table~\ref{tab:main_result} shows the performances of the proposed method and the baselines under different $K$-shot settings. From the table, we can observe that: (1) The proposed method performs consistently better than all the baseline methods. Specifically, the F1-scores of our model advance previous models by +18.72\%, +18.24\%, +14.84\%, +5.37\% on Weibo, WikiGold, CoNLL03 and MIT-Restaurant respectively on 5-shot setting, which verifies the effectiveness of our approach in exploiting few-shot data. (2) Compared to baselines, our method can achieve comparable performance with less training data. Specifically, our approach achieves an F1-score of 55.21\% on CoNLL03 dataset on 5-shot setting, which is better than the result of TemplateNER on 20-shot setting. 

\begin{figure*}[htp]
    \centering{\includegraphics[width=\linewidth]{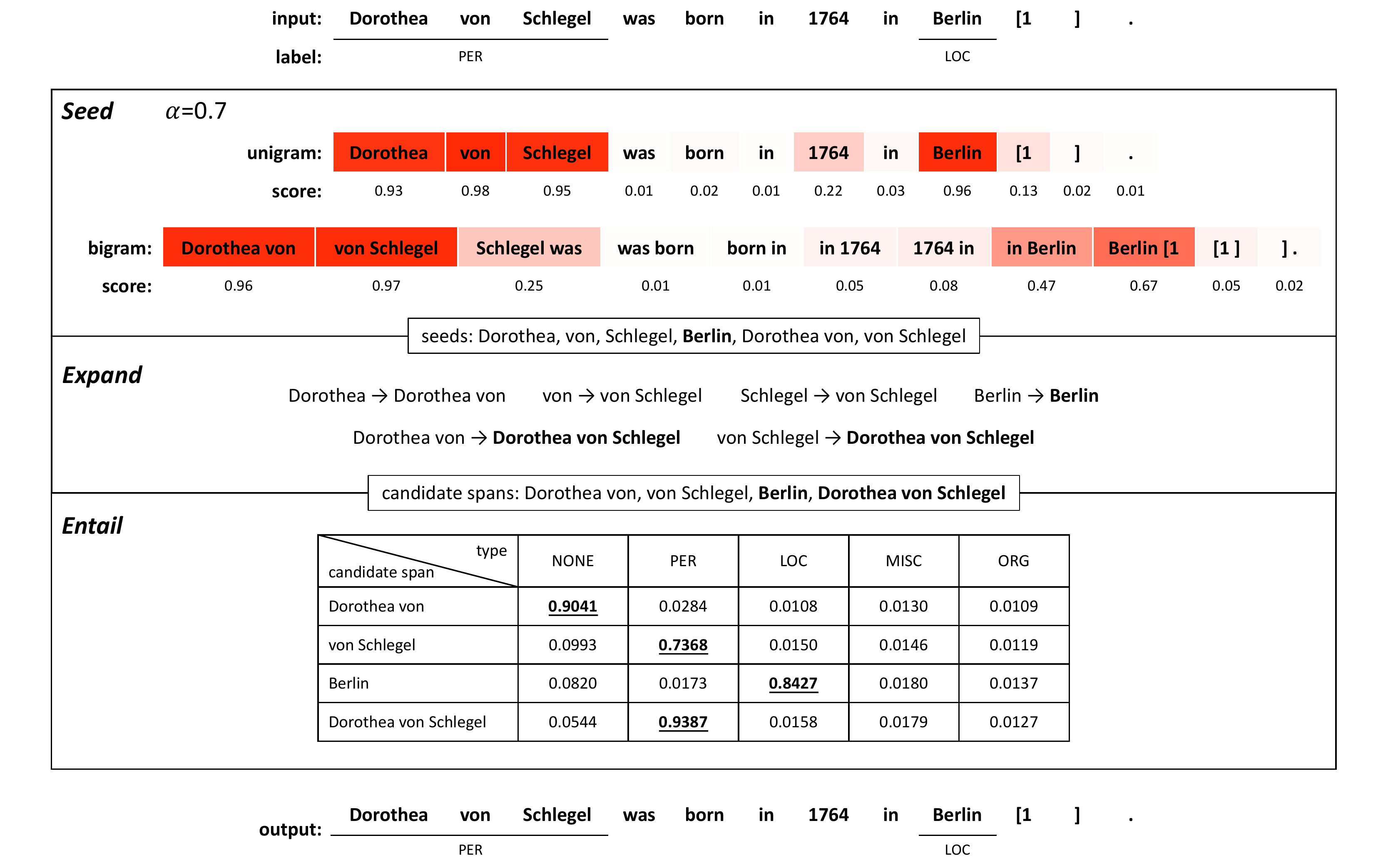}}
    \caption{Case study. The result was predicted by SEE-Few trained on WikiGold, 5-shot setting. }
    \label{fig:visual}
\end{figure*}

\subsection{Ablation Study}

To validate the effectiveness of different components in our approach, we performed ablation experiments on 5-shot setting with a series of variants of SEE-Few. Table~\ref{tab:ablation1} shows the results with the metrics of precision, recall and F1-score, and Table~\ref{tab:ablation2} demonstrates how the number of candidate spans changes in ablations. The variants are as follows:

\textbf{w/o seeding}: removing the seeding module and directly enumerating all unigrams and bigrams as the seeds to expand. With the aid of expanding, this variant reduces 32.25\% unigrams and bigrams on average. The recalls and F1-scores drop by 4.83\% and 2.15\% on average, respectively. The results show that the reduction of candidate span is mainly contributed by the seeding module and this variant suffers from the bias, the tendency to classify named entities as non-entities because of a large number of negative spans. 

\textbf{w/o expanding}: removing the expanding module and directly using the seeds as candidate spans to entail. With the aid of seeding, this variant reduces 83.25\% unigrams and bigrams on average. The recalls and F1-scores drop by 11.74\% and 8.32\% on average, respectively. Without expanding, this variant can not identify the entities whose lengths are greater than 2, achieving worse performance. 

\textbf{w/o entailing}: replacing the entailing module with a multi-class classifier. With the aid of seeding and expanding, this variant reduces 83.5\% unigrams and bigrams on average. The F1-scores drop significantly by 12.17\% on average, while dropping more sharply on datasets with fine-grained entity types (i.e., MIT-Restaurant and Weibo) than on datasets with coarse-grained entity types (i.e., CoNLL03 and WikiGold). The results demonstrate that the improvement comes from exploiting of contextual clues and label knowledge, and the entailing module can better distinguish different entity types than a multi-class classifier. 

\textbf{w/o seed \& exp}: removing both the seeding module and the expanding module in the same way as the w/o seeding and w/o seeding variants. This variant is equivalent to the entailing module classifying all the unigrams and bigrams into the entity categories or the non-entity category. In addition to a significant drop in performance, this variant is time consuming. It is on average 28.50 times slower than the full model on Weibo dataset. 

\textbf{repl IoF with IoU}: using IoU as the ground-truth seed score instead of IoF during seeding, and keeping the threshold $\alpha$s on CoNLL03, MIT-Restaurant, WikiGold and Weibo as 0.5, 0.6, 0.7, 0.7, respectively. The results demonstrate that IoF is a better choice to evaluate the qualities of unigrams and bigrams than IoU. 

All the above experiments show the effectiveness of each component in our approach. Seeding and expanding can significantly reduce the number of candidate spans and alleviate the impact of negative spans in the subsequent span classification stage. The entailing module leverages contextual clues and entity type information benefiting span classification. 

\subsection{Case Study}

Figure~\ref{fig:visual} shows an example of model predictions. We visualize the seed scores and observe that the unigrams and bigrams contained in the ground-truth entities are assigned with higher scores. The threshold $\alpha$ is set to 0.7 in the experiment, so ``Dorothea'', ``von'', ``Schlegel'', ``Berlin'', ``Dorothea von'' and ``von Schlegel'', totally 6 spans, are selected as seeds to expand. Among them, ``Berlin'' already hits the entity exactly, ``Dorothea von'' and ``von Schlegel'' are both expanded to a ground-truth entity ``Dorothea von Schlegel''. Other seeds are not expanded to the ground-truth entities, but do not lead an error in the final output, attributed to the success of the entailing module in determining ``Dorothea von'' is not an entity, and assigning a higher score to ``Dorothea von Schlegel'' with {\tt PER} type than another candidate span (i.e., ``von Schlegel'') overlapping with ``Dorothea von Schlegel''. Considering that in a data-scarce scenario where error propagation is inevitable, our approach can still mitigate the impact of error propagation to a certain extent, which demonstrates the superiority of the proposed paradigm. 

\section{Conclusion}

In this work, we propose a novel multi-task (Seed, Expand and Entail) learning framework, SEE-Few, for Few-shot NER without using source domain data. 
The seeding and expanding modules are responsible for providing as accurate candidate spans as possible for the entailing module. 
The entailing module reformulates span classification as a textual entailment task, leveraging both the contextual clues and entity type information. 
All the three modules share the same text encoder and are jointly learned. 
To investigate the effectiveness of the proposed method, extensive experiments are conducted under the training-from-scratch few-shot setting. 
The proposed method outperforms other state-of-the-art few-shot NER methods by a large margin. 
For future work, we will combine the framework with contrastive learning to effectively make use of limited data and further enhance the performance of few-shot NER. 

\section*{Acknowledgements}

The authors would like to thank the anonymous reviewers for the insightful comments. This work was funded by the National Natural Science Foundation of China (62176053). 

\bibliography{anthology,custom}
\bibliographystyle{acl_natbib}

\appendix

\section{Entity Types and Templates}
\label{app:template}

\begin{CJK}{UTF8}{gbsn}
\begin{table}[ht]
\centering

\resizebox{\linewidth}{!}{
\begin{tabular}{c|c|l}
\hline
\hline
\textbf{Dataset} & \textbf{Entity Type} & \textbf{Template} \\
\hline
\multirow{4}{*}{CoNLL03}
& PER & is a person. \\ 
& LOC & is a location. \\
& MISC & is a miscellaneous entity. \\
& ORG & is an organization. \\
\hline
\multirow{8}{*}{\makecell[c]{ MIT \\ Restaurant}}
& Hours & is a time. \\ 
& Rating & is the rating. \\
& Amenity & is an amenity. \\
& Price & is the price. \\
& Dish & is a dish. \\ 
& Location & is a location. \\
& Cuisine & is is a cuisine. \\
& Restaurant\_Name & is a restaurant name. \\
\hline
\multirow{4}{*}{WikiGold}
& PER & is a person. \\ 
& LOC & is a location. \\
& MISC & is a miscellaneous entity. \\
& ORG & is an organization. \\
\hline
\multirow{8}{*}{Weibo}
& GPE.NAM & 是城市、国家的特指。 \\ 
& GPE.NOM & 是城市、国家的泛指。 \\
& LOC.NAM & 是地名的特指。 \\
& LOC.NOM & 是地名的泛指。 \\
& ORG.NAM & 是组织名的特指。 \\ 
& ORG.NOM & 是组织名的泛指。 \\
& PER.NAM & 是人名的特指。 \\
& PER.NOM & 是人名的泛指。 \\

\hline
\hline
\end{tabular}}
\caption{Entity types and their corresponding natural language templates.}
\label{tab:template}
\end{table}
\end{CJK}

\begin{CJK}{UTF8}{gbsn}
In the span classification stage, the entailing module reformulates span classification as a textual entailment task to leverage contextual clues and entity type information. Table~\ref{tab:template} shows the entity types in each dataset and corresponding natural language templates we use in our experiments. Besides, for English datasets, we use ``is not an entity.'' as the template of the non-entity type. For Chinese datasets, we use ``不是命名实体。'' as the template of the non-entity type. 
\end{CJK}

\end{document}